%% file: main.tex
\documentclass[10pt,twocolumn,letterpaper]{article}

\usepackage[pagenumbers]{iccv} 


\definecolor{iccvblue}{rgb}{0.21,0.49,0.74}
\usepackage[pagebackref,breaklinks,colorlinks,allcolors=iccvblue]{hyperref}

\def\paperID{13425} 
\def\confName{ICCV}
\def\confYear{2025}

\title{\rat: Boosting Misclassification Detection Ability without Extra Data}

\author{Ge Yan\\
UC San Diego\\
{\tt geyan@ucsd.edu}
\and
Tsui-Wei Weng\\
UC San Diego \\
{\tt lweng@ucsd.edu}
}
\input{packages}
\input{symbols}
\begin{document}
\crefname{algocf}{Alg.}{Algs.}
\Crefname{algocf}{Alg.}{Algs.}
\maketitle

\begin{abstract}
As deep neural networks(DNN) become increasingly prevalent, particularly in high-stakes areas such as autonomous driving and healthcare, the ability to detect incorrect predictions of models and intervene accordingly becomes crucial for safety. In this work, we investigate the detection of misclassified inputs for image classification models from the lens of \textit{adversarial perturbation}: we propose to use robust radius (a.k.a. input-space margin) as a confidence metric and design two efficient estimation algorithms, \RobustRadBS~and \RobustRadFast, for misclassification detection. Furthermore, we design a training method called Radius Aware Training (\rat) to boost models' ability to identify mistakes. Extensive experiments show our method could achieve up to 29.3\% reduction on AURC and 21.62\% reduction in FPR@95TPR, compared with previous methods\footnote{Our code will be available at \textsf{\href{https://github.com/Trustworthy-ML-Lab/RAT_MisD}{https://github.com/Trustworthy-ML-Lab/RAT\_MisD}.}}.

  \end{abstract}
\section{Introduction}
\input{sections/introduction}
\section{Background and related works}
\input{sections/background}
\section{Robust radius for misclassification detection}
\input{sections/method}
\section{Experiments}
\input{sections/experiments}
\section{Conclusion and limitations}
\input{sections/conclusion}

\clearpage
\newpage
{
    \small
    \bibliographystyle{ieeenat_fullname}
    \bibliography{ref}
}

\appendix
\onecolumn

\renewcommand{\theequation}{\thesection.\arabic{equation}}
\renewcommand{\thefigure}{\thesection.\arabic{figure}}
\renewcommand{\thetable}{\thesection.\arabic{table}}

\setcounter{equation}{0}
\setcounter{figure}{0}
\setcounter{table}{0}
\input{sections/Appendix/Algorithms}
\input{sections/Appendix/Computation}
\input{sections/Appendix/Experiments}
\input{sections/Appendix/RAT}
\end{document}

%% file: packages.tex
\usepackage{amsmath}
\usepackage{amssymb}
\usepackage{amsthm}
\usepackage{graphicx}
\usepackage{hyperref}
\usepackage{algorithmic}
\usepackage{algorithm2e}
\usepackage{multicol}
\usepackage{multirow}
\usepackage{subcaption}
\usepackage{arydshln}

\RestyleAlgo{ruled}

%% file: symbols.tex
\theoremstyle{plain}

\theoremstyle{definition}

\theoremstyle{remark}

\newcommand{\InputSpace}{\mathcal{X}}
\newcommand{\LabelSpace}{\mathcal{Y}}

\newcommand{\ConfidScore}{C}
\newcommand{\ConfidThres}{\tau}

\newcommand{\FastScore}{r_{\text{fast}}}

\newcommand{\AdvAlgo}{\text{ADV}}

\newcommand{\AdvInput}{\tilde{x}}
\newcommand{\ModelPred}{\hat{y}}
\newcommand{\AdvDirection}{d}
\newcommand{\DirectionCls}{h}
\newcommand{\argmax}{\text{arg\,max}}
\newcommand{\argmin}{\text{arg\,min}}
\newcommand{\mpar}[1]{\left[#1\right]}
\newcommand{\UpperSearch}{L_{\text{upper}}}
\newcommand{\maxiter}{\text{max\_iter}}
\newcommand{\budget}{L_{\text{max}}}
\newcommand{\RobustRad}{\textsf{RR}}
\newcommand{\RobustRadFast}{\textsf{{\small RR-Fast}}}

\newcommand{\rat}{RAT}
\newcommand{\RobustRadBS}{\textsf{{\small RR-BS}}}

%% file: sections/introduction.tex
\label{sec:intro}
Deep neural networks are becoming more prevalent in various applications, including many safety-critical aspects, e.g. autonomous driving~\cite{janai2020autodrive,yurtsever2020survey}. This trend leads to increasing interest in understanding the uncertainty in deep networks. One important task in this area is misclassification detection (MisD)~\cite{granese2021doctor,hendrycks2016MSR,zhu2023openmix}, where the target is to detect the potential error of classification models so that the users can decide whether to trust the model prediction or intervene for potential misclassified examples. 

In order to detect misclassified examples, one key component is confidence score, which measures how confident the model is in a decision. For example, a natural confidence score is the softmax probability prediction from the model, which is known as Max Softmax Response (MSR)~\cite{hendrycks2016MSR}. Unfortunately, it has been shown that MSR suffers from the overconfidence problem~\cite{nguyen2015overconfident}. That is to say, the model may assign a high probability even for erroneous predictions, which hurts the misclassification detection performance when using MSR. Thus, there has been a line of research focusing on designing novel confidence measurements to improve the performance of MSR\cite{gomes2023RelU,granese2021doctor,liang2017ODIN}. 
Despite recent developments, there are still limitations in those confidence scores, e.g. they introduced multiple hyperparameters which need to be tuned on an independent validation set, introducing extra data and computation requirements. 

In this work, we start by developing a better confidence score for misclassification detection, which has two major benefits: (1) has fewer hyperparameters to alleviate the requirement on additional data and computation in hyperparameter tuning, and (2) is more sensitive for detecting misclassified examples. We choose the \textit{robust radius}, also known as input-space margin, which measures the distance of the input to the decision boundary. Robust radius is well studied in traditional statistical learning theory for models like SVM, which is related to the generalization performance. It's also closely related to the adversarial attack~\cite{madry2017pgd} literature where it could be used to detect non-robust examples. While in this work, we propose to use robust radius as a confidence metric for \textbf{misclassification detection} (MisD), which has not been well understood for modern neural networks. In our experiments (\cref{fig:Overview}), we found robust radius servers as a strong metric for detecting wrong classifications: \textit{Misclassified examples are distributed much closer to the decision boundary.}

\begin{figure*}[ht]
    \centering
    \includegraphics[width=\linewidth]{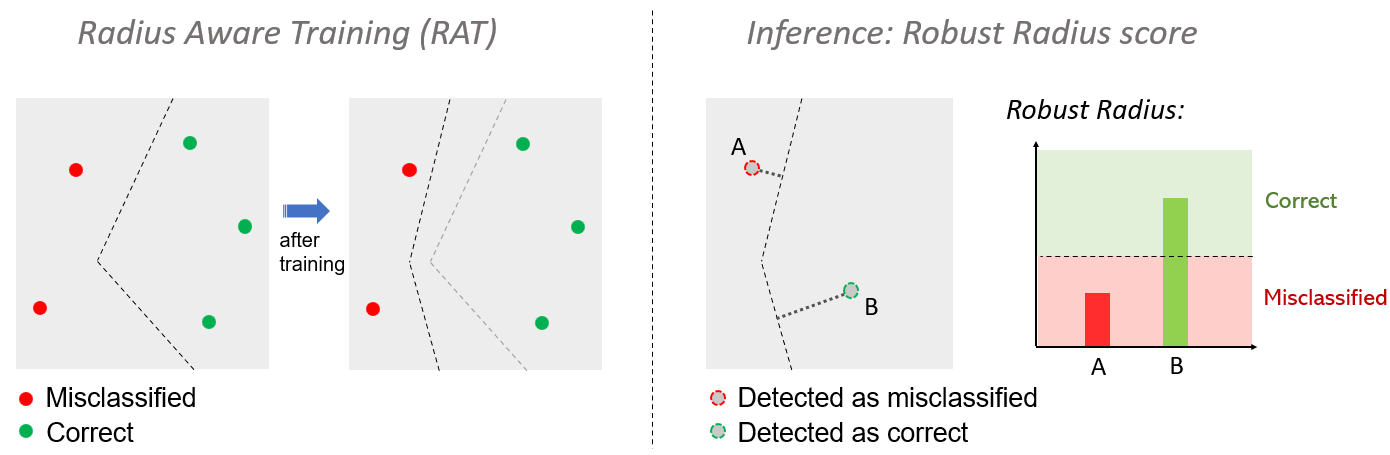}
    \caption{Overview of our method: (Left) In Radius Aware Training, our goal is to make misclassified inputs closer to the boundary, while correct ones further. This helps model distinguish correct and wrong examples. (Right) During inference, we calculate robust radius of each input and use it as a confidence score for detecting potential misclassified inputs. }
    \label{fig:overview-new}
\end{figure*}

Furthermore, we design a training method called Radius Aware Training (\rat) to boost model's ability to detect misclassified inputs. Our key idea is to encourage model to pull misclassified examples closer to boundary while push correct examples away from that. We are inspired by adversarial training \cite{madry2017pgd} and design a specific inner optimization objective to achieve this goal. Experiments show that \rat largely improve the MisD performance of most models and datasets.

Our contributions are summarized below:
\begin{itemize}
    \item We introduce robust radius into the field of Misclassification Detection (MisD) as a confidence score, which is shown to be competitive across multiple benchmarks. Additionally, we design two computation-efficient algorithms: \RobustRadBS~and \RobustRadFast~for efficient robust radius estimation for MisD task. 
    \item Inspired by the importance of robust radius in MisD, we further design a training method called Radius Aware Training (\rat). The key idea is applies designed perturbation on input examples to encourage wrong and correct inputs to be more separated in robust radius. We show this training strategy enhances model's ability to detect misclassification without performance loss. Compared with previous approach, our \rat~  has a key advantage that it \textbf{does not need extra data}. 
    \item We conduct extensive empirical study on our method and several baselines on the task of misclassification detection. Results show that our method outperforms the baselines over most settings, achieving up to 29.3\% reduction on AURC and 21.62\% reduction in FPR@95TPR.
\end{itemize}

%% file: sections/background.tex
\paragraph{Misclassification Detection}Denote $\InputSpace \subseteq \mathbb{R}^d$ as the input space and $\LabelSpace = \{1, 2, \cdots, K\} \triangleq [K]$ as the label space. $f$ is a pretrained classifier with logits $f_i(x), i\in \LabelSpace$. Given an unseen input $x \in \InputSpace$ with ground-truth label $y \in \LabelSpace$, the model predicts the most likely class for $x$ as: $\ModelPred(x) = \argmax_i f_i(x)$. The goal of Misclassification Detection (MisD) is to detect misclassified examples based on model prediction, i.e. $\ModelPred(x) \neq y$.

Conventional approaches for misclassification detection is to utilize the confidence score function to measure prediction confidence and warn users about low-confidence predictions for potential misclassifications. The procedure could be summarized as the following: 
\begin{enumerate}
    \item Define a confidence score function $\ConfidScore: \InputSpace \rightarrow \mathbb{R}$ which measures the confidence of classifier for given input. 
    \item Detect misclassified input via a decision function $g$ for some input $x$ by 
    \begin{equation}
        g(x) = 
    \begin{cases}
        \text{correctly classified, if $\ConfidScore(x) \geq \ConfidThres$;}\\
        \text{misclassified, otherwise.}
    \end{cases}
    \label{eq:DecisionRuleMisDBase}
    \end{equation}
    Here, $\ConfidThres \in \mathbb{R}^+$  is a pre-defined threshold. 
\end{enumerate}

\begin{figure*}[t!]
\begin{subfigure}[b]{0.33\textwidth}
    \centering
    \includegraphics[scale=0.33]{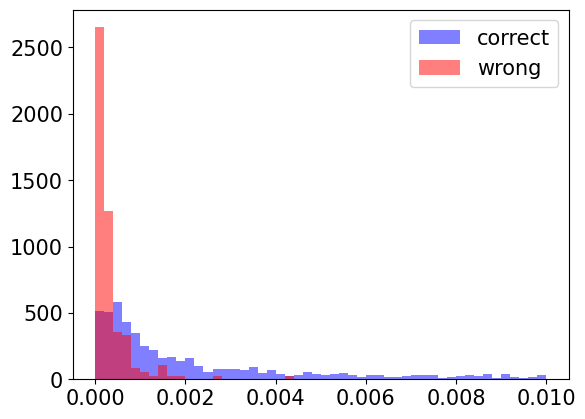}
    \caption{WideResNet}
\end{subfigure}
\hfill
\begin{subfigure}[b]{0.33\textwidth}
    \centering
    \includegraphics[scale=0.33]{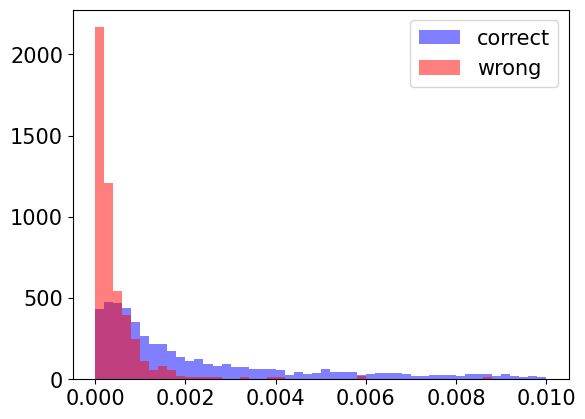}
    \caption{DenseNet}
\end{subfigure}
\hfill
\begin{subfigure}[b]{0.33\textwidth}
    \centering
    \includegraphics[scale=0.33]{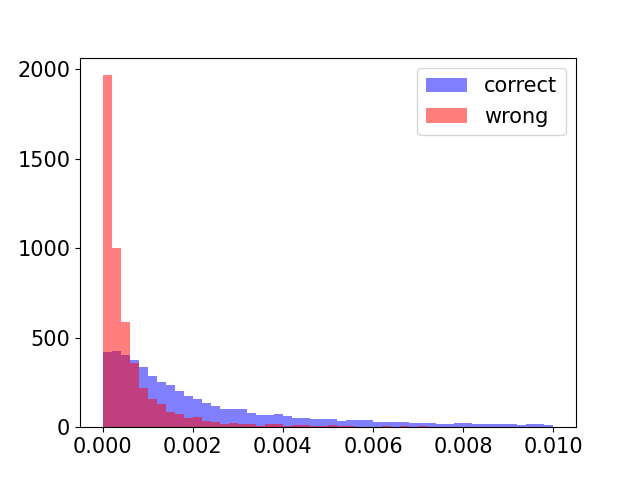}
    \caption{ViT}
\end{subfigure}
\caption{Distribution of robust radius for correctly and wrongly classified examples for different network architectures. For WideResNet\cite{zagoruyko2016wrn} and  DenseNet\cite{huang2017densely} we use the CIFAR10 validation dataset. For ViT\cite{dosovitskiy2020vit} we use the ImageNet validation set. As shown in the figure, for all three architectures, the distribution of robust radius between correctly classified inputs and misclassified inputs clearly differ: misclassified sample (in red color)
generally have smaller robust radius.}
\label{fig:Overview}
\end{figure*}
\paragraph{Confidence score}
Confidence score $\ConfidScore(x)$ is the key component of MisD, which measures (relative) confidence in model predictions. A simple baseline is the Max Softmax Response (MSR)\cite{hendrycks2016MSR}, which uses the softmax probability of the predicted class as the confidence of the prediction. Later, more confidence scores are proposed to improve the performance. \citet{liang2017ODIN} proposed to apply temperature scaling on softmax probability as well as a gradient-based input pre-processing method to boost performance. \citet{granese2021doctor} built a mathematical framework for misclassification detection and designed a novel confidence score called DOCTOR. \citet{gomes2023RelU} suggested using a data-driven method to estimate uncertainty in the decision and called this RelU.

One drawback of these methods is that they introduced multiple hyperparameters: All these methods~\cite{liang2017ODIN, granese2021doctor,gomes2023RelU}  require at least two hyperparameters: softmax temperature $T$ and input perturbation magnitude $\epsilon$. Additionally, those scores are also sensitive to hyperparameter choice, as we shown in \cref{fig:TempStudy}. Consequently, users need to tune these hyperparameters by grid-search on a separate validation set to achieve good performance, which imposes an extra requirement on data and computation. 

In this work, we explore a novel confidence score, robust radius, which we find has several advantages: It reduces the cost of hyperparameter tuning and outperforms the baseline in many misclassification detection benchmarks.

\paragraph{Adversarial attacks and misclassifications}
\citet{goodfellow2014fgsm} showed that deep networks are prone to specifically crafted small perturbations, which could fool the model into making wrong predictions. This process is known as adversarial attack. Common adversarial attack algorithms include FGSM~\cite{goodfellow2014fgsm}, PGD~\cite{madry2017pgd} and AutoAttack~\cite{croce2020autoattack}. In this work, for the purpose of efficient computation, we adopt FGSM to calculate robust radius, which is efficient and effective for misclassification detection tasks.

 Previously, several works study the relation between adversarial attacks and model misclassification. \citet{schwinn2023exploring} studied the distribution of images that are misclassified under adversarial attack and proposed a novel loss for adversarial attack. \citet{Wang2020Improving} studied the importance of misclassified examples in adversarial training and proposed a novel adversarial training scheme. \citet{haffar2021explaining} used adversarial examples to explain the failure cases of deep neural networks. Most of the previous works focus on improving adversarial attack/defense using misclassified examples. In contrast, in this work, we study how sensitivity to adversarial perturbation could be used to \textit{detect misclassified examples}. 


\paragraph{Training methods for misclassification detection}
Recently, \citet{zhu2023openmix} proposed to improve the misclassification detection ability by incorporating a new training scheme called OpenMix. OpenMix utilizes outlier examples to generate uncertain pseudo-examples, and the model learns to reject them during training. The major limitation of this method is that it requires extra outlier data to generate pseudo-exmaples: For example, in CIFAR10 dataset they used 300K RandImages~\cite{hendrycks2018deep} dataset which is $6\times$ larger than the original training set. This demand on extra data limits their usage, especially when high-quality data is expensive to access. On the other hand, our \rat~introduced in \cref{sec:rat} \textbf{does not need any extra data}, providing a data-efficient way for improving MisD performance. 

%% file: sections/method.tex
In this section, we first briefly introduce the definition of robust radius (a.k.a. input-space margin). Then, we show robust radius is a strong confidence metric for MisD, which is not explored in this area before. Further, we design two computationally efficient methods, \RobustRadBS~ and \RobustRadFast, to estimate the robust radius for detecting misclassified inputs.

\subsection{Definition}
The idea of robust radius has been introduced for long in the literature of statistical learning. For example, in Support Vector Machine (SVM)~\cite{hearst1998support}, the goal is to maximize the distance of inputs to the decision plane, which is exactly the robust radius. For modern neural network, the robust radius is usually studied in the adversarial attack literature, where it measures the scale of adversarial perturbation required to alter the decision of the input. 
The robust radius could be formally defined as:
\begin{equation}
\begin{aligned}
    \RobustRad = \argmin \quad & {\|\delta\|_{\infty}} \\ 
    \text{s.t.} \quad &  \ModelPred(x + \delta) \neq \ModelPred(x).
\end{aligned}
\end{equation}
Here, $\ModelPred(x) = \argmax_i f_i(x)$ is the model prediction, and we use the $L_\infty$ norm which is frequently studied in the literature of adversarial attack~\cite{szegedy2013adv}. Intuitively, it measures the minimum perturbation required to alter the decision of the model.


\paragraph{Robust radius as a confidence measure}
Previous works on robust radius are mostly studied in the adversarial robustness community with main focus on evaluating or enhancing the robustness of a given neural network via attacks (i.e. crafting adversarial perturbation), robustness verification (i.e. providing robustness certificate), and defense (i.e. improving robustness of DNNs through training or detecting adversarial threats). 
%
However, in this work, our research focus differs from prior works -- our goal is to \textit{use robust radius as a confidence metric for \textbf{misclassification detection} task (MisD).} 
%

The motivation of using robust radius as a confidence measure roots from a natural intuition: \textit{the inputs that located near the decision boundary (i.e. have small robust radius), should have less confidence than those far from the boundary (i.e. have large robust radius)}. To verify this intuition, we compare the robust radius of correct and misclassified inputs on different network architectures and datasets and show the results in \cref{fig:Overview}. It can be seen that experiments support our intuition, showing that misclassified examples have smaller robust radius than correct ones. Further, we conduct more quantitative studies on \cref{sec:exp} to compare robust radius with other common confidence metrics, showing that robust radius outperforms other baselines over many benchmarks. 

\subsection{Fast approximation of robust radius}
Exactly computing the robust radius for modern deep neural networks is known to be intractable~\cite{katz2017reluplex,jordan2020exactly}. As a result, various approximation methods are developed, e.g., DeepFool~\cite{moosavi2016deepfool} and FAB~\cite{croce2020minimally}. 
These algorithms could be used out-of-the-box for misclassification detection. However, since they are originally developed for benchmarking the adversarial robustness of different models, they focus on \textit{accurately calculating robust radius} at the cost of heavy computational overhead. In experiments, we found that the misclassification detection performance is less sensitive to the accuracy of radius estimation, while \textit{detection efficiency} may be more important for users in practice, as it helps reduce inference latency. Thus, we design the following two algorithms, \RobustRadBS~and \RobustRadFast, to provide \textit{efficient} radius estimation for MisD task. These algorithms are easy to implement, fast, and provide competitive results in MisD as we show in \cref{tab:RRCompare}.
\paragraph{\RobustRadBS} The \RobustRadBS~ is a simple adaptation of the FGSM adversarial attack. We use FGSM algorithm to decide whether model decision could be altered at a given budget $r$, and use a binary search to find the smallest radius that changes the model's prediction. The full algorithm is provided in \cref{alg:RR-BS}.

\paragraph{\RobustRadFast} The motivation is that although \RobustRadBS~only needs a single backward step to get gradient and set the perturbation direction, it still requires multiple forward propagations due to the binary search step.
To further reduce the computational cost of \RobustRadBS, we propose to use linear approximation to avoid the expensive binary search, allowing even faster speed. The key idea is summarized below:
\begin{enumerate}
    \item \textbf{Select perturbation direction:} In FGSM, the search direction of adversarial perturbation is determined by $\text{sign}(\nabla_x l(x, y))$, regardless of $\|\delta\|$. This inspires us to choose a direction $\AdvDirection$ and search perturbation along this direction. Here, we follow the FGSM and choose the sign of the gradient of cross-entropy loss ($l = CE$) as the perturbation direction. The difference is, instead of using the ground truth label $y$, we plug in the predicted label $\ModelPred$:
    \begin{equation}
        d(x) = \text{sign}(\nabla_x \mpar{CE(x, \ModelPred)}),
        \label{eq:FGSMDirection}
    \end{equation}
    where $CE(\cdot, \cdot)$ denotes the cross-entropy loss.
    \item \textbf{Linear approximation:} After choosing a direction $\AdvDirection$, the classifier could then be parameterized by $t$:
    \begin{equation}
        \DirectionCls_i(t) = f_i(x + t\AdvDirection), i = 1, \cdots, K,
    \end{equation}
    ,where $K$ is the number of classes. Note that from \cref{eq:FGSMDirection}, $\|d\|_\infty = 1$. Now the problem  reduces to searching the smallest $t$ such that $\argmax_i \DirectionCls_i(t) \neq \argmax_i \DirectionCls_i(0) = \argmax_i f(x) = \ModelPred(x)$, i.e. finding $t$ such that the prediction at radius $t$ is different from prediction at radius $0$ (i.e. the original prediction). To derive a fast approximation, We perform a linear approximation on $\DirectionCls_i(t)$: 
    \begin{equation}
        \DirectionCls_i(t) \approx \DirectionCls_i(0) + t\DirectionCls_i^\prime(0).
    \end{equation} 
     Calculating all $K$ derivatives $\DirectionCls_i^\prime(0)$ via back-propagation takes $K$ backward pass. To reduce computation, we use a finite difference approximation:
    \begin{equation}
        \DirectionCls^\prime(0) \approx \frac{\DirectionCls(\alpha) - \DirectionCls(0)}{\alpha},
    \end{equation}
    where $\alpha$ is a small constant. In experiments, we take $\alpha = 0.01$.
    \item \textbf{Solve optimization problem:} Given the linear approximation, we could derive an optimization problem:
    \begin{equation}
        \begin{aligned}
            \text{minimize}\quad & t \\
            \text{s.t.}\quad & t \geq 0 \\
            & \argmax_i \: [\DirectionCls_i(0) + t\DirectionCls^\prime_i(0)] \neq \ModelPred(x)
        \end{aligned}
        \label{eq:FastOpt}
    \end{equation}
    This optimization problem has a closed-form solution: Let 
    \begin{equation}
        q_i = \begin{cases}
            \infty, & \frac{\DirectionCls_{\ModelPred}(0) - \DirectionCls_i(0)}{\DirectionCls_{\ModelPred}^\prime(0) - \DirectionCls_i^\prime(0)} < 0;\\
            \frac{\DirectionCls_{\ModelPred}(0) - \DirectionCls_i(0)}{\DirectionCls_{\ModelPred}^\prime(0) - \DirectionCls_i^\prime(0)}, & otherwise.
        \end{cases}
        \label{eq:FastSolution}
    \end{equation}
    Then,
    $t^* = \min_{i \neq \ModelPred}\:q_i$ is the solution of \cref{eq:FastOpt} and we get the estimated minimal perturbation $\FastScore = t^*\|d\|_\infty = \min_{i \neq \ModelPred}\:q_i$.
\end{enumerate}
We summarize these steps in \cref{alg:RRfast} in \cref{app:algo}.

\paragraph{Experiments} We compare \RobustRadBS~and \RobustRadFast~with standard methods including FAB and DeepFool in Sec~\ref{sec:exp} (e.g.\cref{tab:RRCompare}). Results show our method achieve comparable MisD performance while improves the speed by $3.37\times$ for \RobustRadBS~ and $46.5\times$ for \RobustRadFast.
\section{Radius Aware Training: Improving detection performance without extra data}
\label{sec:rat}
\paragraph{Motivation} In previous section, we show that show correct and misclassified examples could be distinguished via different robust radius. Inspired by this, we propose to further boost this distinction by incorporating robust radius into the training pipeline. A highly related, well-studied literature is adversarial training \cite{madry2017pgd,shafahi2019adversarial}, which exposes the model to adversarial examples during training to improve its robustness to adversarial noises.
Common adversarial training objective could be formally written as below:
\begin{equation}
    \min_{\Theta} \max_{\|x^\prime - x_0\|_\infty \leq \epsilon} l(\Theta; x^\prime),
    \label{eq:ObjAT}
\end{equation}
where $\Theta$ is trainable model parameters. Usually, adversarial attack algorithms (e.g. FGSM~\cite{goodfellow2014fgsm}, PGD~\cite{madry2017pgd}) are applied to solve the inner maximization problem to generate adversarial examples, then models are trained on those adversarial examples with standard loss. 

A intuitive understanding of adversarial training is, it punishes inputs that are close to decision boundary even they are correct (non-robust inputs), as their adversarial examples will be wrongly classified. Therefore, this approach should be able to make model predictions consistent under small perturbations, which increases the robustness radius. We conducted an experiment to evaluate the robust radius on Adversarially-Trained (AT) models in \cref{tab:ATRadius} which verifies this. However, the misclassification detection performance (measured by the AUROC) does not improve after AT. The key reason is: the robust radius of both correct and misclassified examples is extended, making it difficult to distinguish two classes of inputs. 

\input{tables/at-radius}

Since adversarial training provides a way to increase the robust radius, we ask the next question: \textit{can we find a way to reduce the radius of wrong examples}? To achieve this, we modify the regular adversarial training objective, changing the inner maximization to minimization:
\begin{equation}
    \min_{\Theta} \min_{\|x^\prime - x_0\|_\infty \leq \epsilon} l(\Theta; x^\prime).
    \label{eq:ObjReversedAT}
\end{equation}
This objective could be regarded as 
a reversed version adv-training (denoted as ReverseAT in \cref{tab:ATRadius}): it allows model to make ``mild" mistake, as long as the input is close to the correct decision region, because the perturbed inputs will then get correctly classified. As shown in \cref{tab:ATRadius}, as we expected, training model with this reversed AT objective reduces the robust radius of both correct and misclassified examples. But it does not improve MisD performance due to the same reason of AT.

\paragraph{Radius Aware Training (\rat)} Inspired by the experiments above, we design a novel method to improve the misclassification detection ability of the model by combining these two objectives. The idea is, since our goal is to make wrong inputs \textit{closer} to the decision boundary and make correct inputs \textit{further}, we should apply regular AT objective on correct training samples, while apply the reversed AT objective on wrong samples. We call this training objective Radius Aware Training (\rat) objective. Formally, the training objective could be written as:
\begin{equation}
    L_{\text{RAT}}(\Theta; x, y) = \begin{cases}
        \min_{\|x^\prime - x_0\|_\infty \leq \epsilon} l(\Theta; x^\prime) & \text{if } \ModelPred(x) \neq y;\\
        \max_{\|x^\prime - x_0\|_\infty \leq \epsilon} l(\Theta; x^\prime) & \text{if } \ModelPred(x) = y.
    \end{cases} 
    \label{eq:ObjRAT}
\end{equation}\
This \rat~ training objective could be optimized using any adversarial attack algorithm. In this work, we use the simple FGSM~\cite{goodfellow2014fgsm} attack as an illustration of concept. 
In training, we combine \rat~ objective with standard cross-entropy loss, giving the final objective:
\begin{equation}
    L(\Theta; x, y) = L_{\text{CE}}(\Theta;x, y) + L_{\text{RAT}}(\Theta; x, y).
\end{equation}
We show an illustration of our method in \cref{fig:method_comp}. Intuitively, we encourage the model to increase the robust radius for those examples that could be correctly classified, while alleviating the training difficulty, allowing the model to make mild mistakes as long as it's not far from the correct decision region. 

%% file: tables/at-radius.tex
\begin{table}[b!]
\centering
\scalebox{0.94}{
\begin{tabular}{ccp{2cm}p{2cm}}
\toprule
 Method & AUROC $\uparrow$ &  Radius$\times 10^4$ (correct) &Radius$\times 10^4$ (wrong)\\
\midrule
Standard & 
0.8293  & 8.91 & 1.67 \\
 AT & 0.8249  & 33.38 & 4.91 \\
 ReverseAT  & 0.8246 & 2.49 & 0.45 \\
\bottomrule
\end{tabular}
}
\caption{Comparison of median robust radius for correct and misclassified examples. The base model studied here is ResNet18~\cite{he2016deep} and attack budget for AT and ReverseAT is 0.001. The radius norm here is $L_\infty$ and we calculate the median of all inputs. From the table, we could see that AT increases the robust radius while ReversedAT decreases it. However, since those methods treat correct and misclassified inputs in the same way, the difference between correct and wrong inputs (measured by AUROC) does not improve. }
\label{tab:ATRadius}
\end{table}

%% file: sections/experiments.tex
\label{sec:exp}
\subsection{Setup}
\input{tables/main_results_new}

In this section, we conduct extensive empirical study on our algorithms and several baselines. More experimental details and results are presented in \cref{app:exp}.
\paragraph{Dataset and architecture.}
We conduct our experiments on CIFAR10, CIFAR100~\cite{krizhevsky2009cifar} and ImageNet~\cite{Deng2009ImageNet}. For each dataset, we randomly sample a subset that contains $20\%$ of test examples as the validation set for tuning hyper-parameters. We test with two popular architectures: ResNet \cite{he2016deep} and WideResNet~\cite{zagoruyko2016wrn}. More specifically, we use ResNet110 and WRN-28-10 on CIFAR10 and CIFAR100, ResNet50 for ImageNet. 
\paragraph{Evaluation metrics}
We report three metrics: AURC, FPR95 and AUROC. We choose AURC as the major metric for evaluation and use this to tune the hyper-parameters.
\begin{enumerate}
    \item \textbf{AURC}\cite{geifman2017selective}: The area under risk-coverage Curve shows how the error rate changes for different confidence thresholds. 
    \item \textbf{FPR95}: False Positive Rate(FPR), a.k.a. False Rejection Rate in misclassification detection literature, is the probability that a correct example is rejected. True Positive Rate(TPR) is the probability that a misclassified example is rejected. FPR95 measures the FPR when TPR is 95\%.
    \item \textbf{AUROC}\cite{davis2006auc}: The area under the receiver operating characteristic curve depicts the relationship between FPR and TPR under different thresholds. 
\end{enumerate}
\input{tables/rr_compare}

\paragraph{Baselines}
For confidence scores, besides the most popular baseline MSR~\cite{hendrycks2016MSR}, we also compare with several recent developments in confidence scores: DOCTOR~\cite{granese2021doctor}, ODIN~\cite{liang2017ODIN}, and RelU~\cite{gomes2023RelU}. RelU differs from others as it requires additional training data to train an uncertainty estimator. Since it does not need to retrain the base classifier, we still categorize it into a confidence score, and we split half of the validation set for training its uncertainty estimator for fair comparison. For training based methods, beside current SOTA method OpenMix~\cite{zhu2023openmix}, we also compare with mixup-based methods including Mixup~\cite{zhang2017mixup} and RegMixup~\cite{pinto2022using}, as they show competitive performance in previous studies~\cite{zhu2023openmix}. Since OpenMix requires extra data for training, we use 300k RandomImage~\cite{hendrycks2018deep} for OpenMix training on CIFAR10 and CIFAR100, Places365~\cite{zhou2017places} on ImageNet. 

\subsection{Main results}
In this section, we conduct experiments on three vision datasets and derive results in \cref{tab:MainResultsCIFAR10,tab:MainResultsImageNet}, where we run the experiments with 3 different random val-test splits and report the mean and standard deviation. We postpone the CIFAR100 results to \cref{tab:MainResultsCIFAR100} due to space limit. From the results, we could see that (1) Our method consistently outperform all confidence-score based methods over all dataset/architecture combinations. On CIFAR10 dataset, our method achieves up to 39.14\% reduction in FPR95 and 45.98\% reduction in AURC, compared with best confidence-score based method. (2) Among training-based methods, our method is generally competitive with current SOTA training method OpenMix, and outperforms it over multiple benchmarks. Our results are achieved \textbf{without any extra data}, while OpenMix use large-scale extra data to enhance the training. 

\subsection{Ablation study for \rat}
In this section, we conduct an ablation study on \rat. We compare it with (1) standard trained model (2) adversarial training objective in \cref{eq:ObjAT} and (3) reversed adversarial training objective in \cref{eq:ObjReversedAT}. For different AT objective, we both use the FGSM algorithm to train. The results could be found at \cref{tab:ATAblation}, which shows our \rat~performs better than traditional AT objective or the reversed AT objective, suggesting the effectiveness of our method. 
\input{tables/ablation_at}

\subsection{Comparison for different radius calculation methods}
In this section, we discuss the calculation of robust radius. For comparison, besides two traditional algorithm FAB and DeepFool and our \RobustRadBS~and \RobustRadFast, we also add a version of \RobustRadBS~that uses PGD~\cite{madry2017pgd} as the base algorithm. From the results in \cref{tab:RRCompare}, we could see: (1) All robust radius based methods largely outperform the MSR baseline, while among them \RobustRadBS~(PGD) and FAB have best performance (2) Regarding computation speed, \RobustRadFast~ is the fastest, achieving a speed comparable to MSR and $\textbf{46.5}\times$ faster than FAB. Combining these, \RobustRadFast~provides a good balance between detection speed and performance. 

\begin{figure*}[bht!]
\begin{subfigure}[b]{0.48\textwidth}
    \centering
    \includegraphics[scale=0.4]{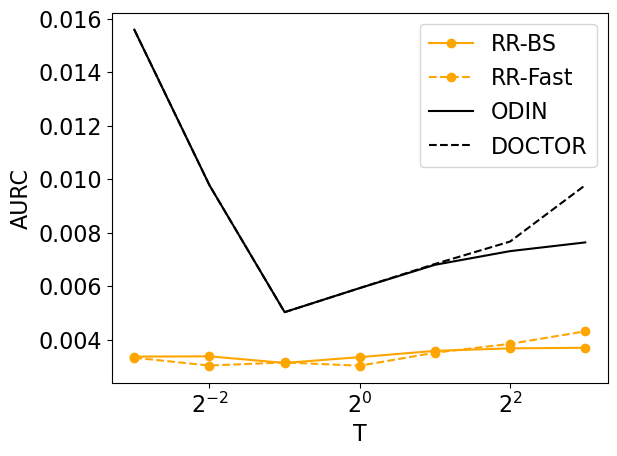}
    \caption{CIFAR10}
\end{subfigure}
\hfill
\begin{subfigure}[b]{0.48\textwidth}
    \centering
    \includegraphics[scale=0.4]{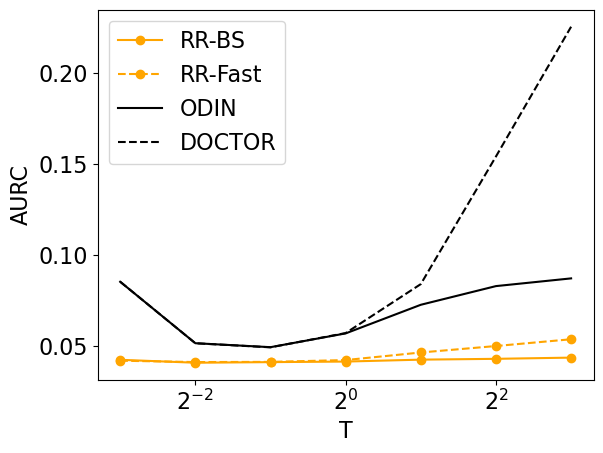}
    \caption{CIFAR100}
\end{subfigure}
\caption{\textbf{AURC (lower is better) vs. temperature $T$ on CIFAR10 and CIFAR100 datasets:} This figure shows the sensitivity of different methods to the hyperparameter $T$. The orange lines denote our methods and the black lines denote two baselines, ODIN and DOCTOR. Compared with the baselines, our methods are less sensitive to the choice of $T$.}
\label{fig:TempStudy}
\end{figure*}
\subsection{Hyper-parameter sensitivity of robust radius}
As we discussed in previous sections, one limitation of the previous confidence scores is they require multiple hyperparameters, including softmax temperature $T$ and perturbation magnitude $\epsilon$.  In practice, in order to choose these hyperparameters, users usually need to conduct a grid search over a validation set, which demands additional data and computation. Using robust radius as a confidence score, however, is much less demanding in choosing hyperparameters for following two reasons: 
\begin{enumerate}
    \item \textbf{Robust radius requires less hyperparameters: }\RobustRadBS~ and \RobustRadFast~ only depend on the temperature $T$.
    \item \textbf{Robust radius  are less sensitive to the choice of hyperparameters:} To understand the sensitivity to $T$, we conduct experiments across different temperatures and compare the AURC. As shown in \cref{fig:TempStudy}, our methods \RobustRadBS~and \RobustRadFast~ are less sensitive to temperature, while DOCTOR and ODIN are much more sensitive. This suggests users could apply our methods without any validation data and hyperparameter tuning and still achieve good performance.  
\end{enumerate} 

\subsection{Detect misclassified examples under input corruption}
\begin{figure}[htb!]
\begin{subfigure}[b]{0.48\textwidth}
    \centering
    \includegraphics[scale=0.2]{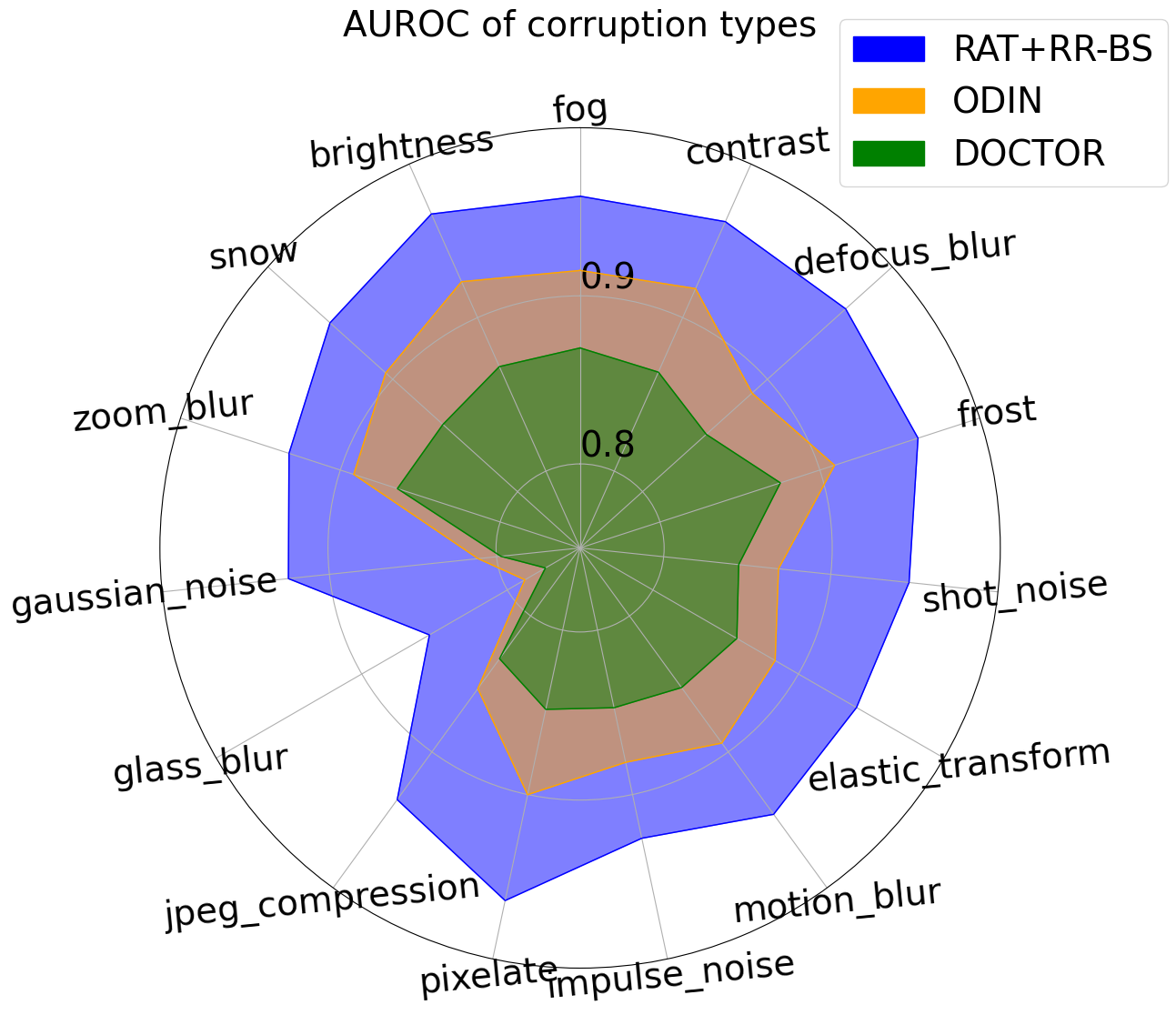}
    \caption{AUROC}
\end{subfigure}
\hfill
\begin{subfigure}[b]{0.48\textwidth}
    \centering
    \includegraphics[scale=0.2]{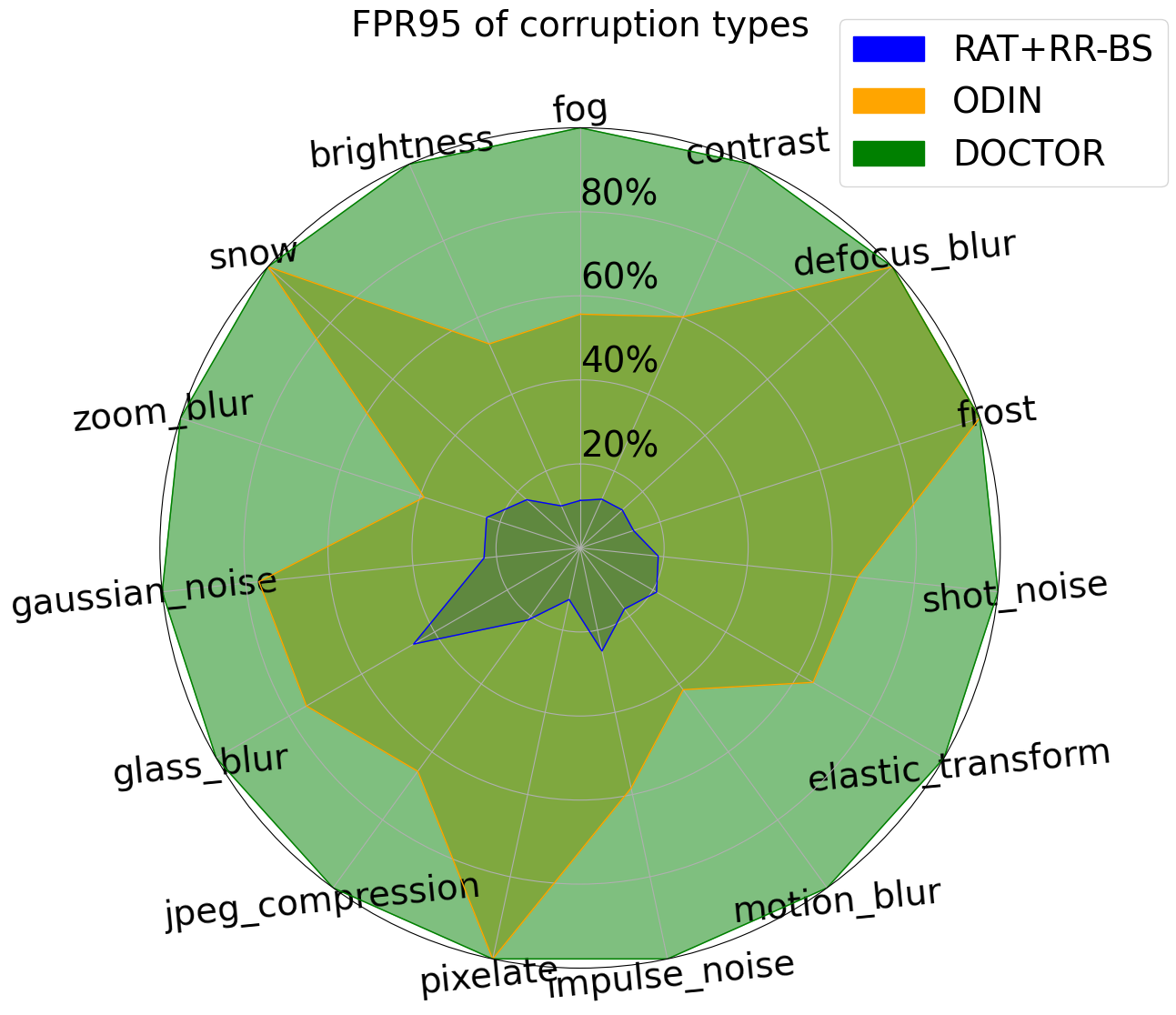}
    \caption{FPR95}
\end{subfigure}
\caption{\textbf{AUROC(Top, higher is better) and FPR95 (Bottom, lower is better) of different corruption types over CIFAR10-C dataset.} As shown in the figure, our method outperforms the baselines over all different corruptions type, showing that our method is still valid under noisy inputs.}
\label{fig:Corrupt}
\end{figure}

In this section, we study the ability of our methods under input corruption. We chose the CIFAR10-C~\cite{hendrycks2019benchmarking} dataset, which contains images under 15 common corruptions. We use a validation set consisting of clean CIFAR10 images and test on the corrupted version of the remaining datasets, which helps us to better understand the impact of the covariate shift at test time. We plot AUROC and FPR95 in \cref{fig:Corrupt}. As the results suggest, our method outperforms the baselines over all different kinds of corruptions, showing our robustness to input corruptions. 

%% file: tables/main_results_new.tex
\begin{table*}[htbp]
\centering
\scalebox{0.9}{
\begin{tabular}{c|c|ccccc}
\toprule
Model &  Method & AUROC $\uparrow$& FPR95 $\downarrow$ & AURC $\times$ 1000 $\downarrow$ & Accuracy $\uparrow$& \# of extra samples$\downarrow$\\
\midrule
\multirow{9}{*}{WRN28} & MSR & 0.8980 $\pm$ 0.0123 & 58.73 $\pm$ 3.15\% & 4.43 $\pm$ 0.45 & 97.55\% & -  \\
 & ODIN & 0.9132 $\pm$ 0.0047 & 52.82 $\pm$ 5.68\% & 3.48 $\pm$ 0.05 & 97.55\% & -   \\
 & DOCTOR & 0.8868 $\pm$ 0.0116 & 90.65 $\pm$ 16.19\% & 5.54 $\pm$ 0.47 & 97.55\% & -  \\
 & RelU & 0.8085 $\pm$ 0.0123 & 79.14 $\pm$ 3.56\% & 8.54 $\pm$ 0.29 & 97.55\% & -  \\
 & OpenMix & \underline{0.9548 $\pm$ 0.0016} & \textbf{12.22 $\pm$ 2.14\%} & \underline{2.01 $\pm$ 0.06}& 97.12\% & 300k \\
 & Mixup & 0.9341 $\pm$ 0.0023 & 27.65 $\pm$ 6.76\% & 3.13 $\pm$ 0.18 & 96.80\% & 0  \\
 & RegMixup & 0.9157 $\pm$ 0.0023 & 31.34 $\pm$ 1.66\% & 3.09 $\pm$ 0.10 & 97.36\% & 0  
\\
\cdashline{2-7}\noalign{\vskip 0.5ex}
 & \rat+\RobustRadBS~\textbf{(Ours)} & \textbf{0.9585 $\pm$ 0.0034} & \underline{13.68 $\pm$ 2.40\%} & \textbf{1.60 $\pm$ 0.19} & \textbf{97.60\%} & 0  \\
 & \rat+MSR \textbf{(Ours)} & 0.9373 $\pm$ 0.0091 & 33.76 $\pm$ 19.56\% & 2.36 $\pm$ 0.41 & \textbf{97.60\%} & 0  \\
 \midrule
\multirow{9}{*}{ResNet110} & MSR & 0.9238 $\pm$ 0.0017 & 38.21 $\pm$ 8.76\% & 5.09 $\pm$ 0.42 & \textbf{96.40\%} & -  \\
 & ODIN & 0.9258 $\pm$ 0.0031 & 54.05 $\pm$ 39.83\% & 4.87 $\pm$ 0.49 & \textbf{96.40\%} & -  \\
 & DOCTOR & 0.9191 $\pm$ 0.0019 & 80.87 $\pm$ 33.14\% & 5.51 $\pm$ 0.19 & \textbf{96.40\%} & -  \\
 & RelU & 0.8954 $\pm$ 0.0045 & 59.35 $\pm$ 3.54\% & 7.02 $\pm$ 0.69 & \textbf{96.40\%} & -  \\
 & OpenMix & 0.9208 $\pm$ 0.0014 & 28.84 $\pm$ 1.28\% & 5.96 $\pm$ 0.13 & 95.39\% & 300k  \\
 & Mixup & 0.8987 $\pm$ 0.0041 & 54.45 $\pm$ 2.49\% & 9.22 $\pm$ 0.52 & 94.46\%& 0  \\
 & RegMixup & 0.9129 $\pm$ 0.0012 & 39.68 $\pm$ 1.31\% & 6.16 $\pm$ 0.17 & 95.38\% & 0  \\
 \cdashline{2-7}\noalign{\vskip 0.5ex}
 & \rat+\RobustRadBS~\textbf{(Ours)}& \textbf{0.9480 $\pm$ 0.0031} & \textbf{16.59 $\pm$ 1.74\%} & \textbf{3.60 $\pm$ 0.39} & \underline{95.68\%} & 0   \\
 & \rat+MSR \textbf{(Ours)}& \underline{0.9412 $\pm$ 0.0039} & \underline{19.90 $\pm$ 4.32\%} & \underline{3.77 $\pm$ 0.22} & \underline{95.68\%} & 0  \\
\bottomrule
\end{tabular}
}
\caption{\textbf{Main results for our methods and baselines on CIFAR10 dataset.} Standard deviation is measures over 3 independent runs. For OpenMix~\cite{zhu2023openmix}, the OOD dataset used is 300K-Random-Image dataset~\cite{hendrycks2018deep}.}
\label{tab:MainResultsCIFAR10}
\end{table*}

\begin{table*}[htbp]
\centering
\scalebox{0.9}{
\begin{tabular}{c|c|ccccc}
\toprule
Model &  Method & AUROC $\uparrow$& FPR95 $\downarrow$ & AURC $\times$ 1000 $\downarrow$ &  Accuracy $\uparrow$ & \# of extra samples$\downarrow$\\\midrule
\multirow{7}{*}{ResNet50} &  MSR & 0.8644 $\pm$ 0.0005 & \underline{45.95 $\pm$ 0.29\%} & 69.69 $\pm$ 0.35 & 76.20\% & -\\
 & ODIN & 0.8644 $\pm$ 0.0005 & \underline{45.95 $\pm$ 0.29\%} & 69.69 $\pm$ 0.35 & 76.20\%& -\\
 & DOCTOR & 0.8641 $\pm$ 0.0010 & 46.09 $\pm$ 0.38\% & 69.79 $\pm$ 0.37 & 76.20\% & -\\
 & RelU & 0.8010 $\pm$ 0.0049 & 59.69 $\pm$ 1.30\% & 91.00$\pm$ 2.34 & 76.20\% & -\\
 & OpenMix & \underline{0.8674 $\pm$ 0.0018} & \textbf{45.38 $\pm$ 0.70\%} & \underline{65.53 $\pm$ 0.42} & \underline{76.98\%} & 1.8M \\
 \cdashline{2-7}\noalign{\vskip 0.5ex}
 & \rat+\RobustRadBS~\textbf{(Ours)} & 0.8465 $\pm$ 0.0004 & 53.10 $\pm$ 0.48\% & 71.85 $\pm$ 0.04 & \textbf{77.29\% }& 0\\
 & \rat+MSR \textbf{(Ours)} & \textbf{0.8679 $\pm$ 0.0007} & 47.37 $\pm$ 0.52\% & \textbf{64.99 $\pm$ 0.22} & \textbf{77.29\%} & 0\\
\bottomrule
\end{tabular}
}
\caption{\textbf{Main results for our methods and baselines on ImageNet dataset.} Standard deviation is measures over 3 independent runs. For OpenMix~\cite{zhu2023openmix}, the OOD dataset used is Places365~\cite{zhou2017places}.}
\label{tab:MainResultsImageNet}
\end{table*}

%% file: tables/rr_compare.tex
\begin{table*}[htbp]
\centering
\scalebox{0.95}{
\begin{tabular}{lcccc}
\toprule
  Method & AUROC $\uparrow$& FPR95 $\downarrow$ & AURC $\times$ 1000 $\downarrow$ & Throughput (images/s)\\
\midrule
  MSR & 0.9373 $\pm$ 0.0091 & 33.76 $\pm$ 19.56\% & 2.36 $\pm$ 0.41 & 952.4
  \vspace{1pt}\\
\cdashline{1-5}\noalign{\vskip 0.5ex}
  FAB & 0.9634 $\pm$ 0.0013 & 14.25 $\pm$ 1.44\% & 1.27 $\pm$ 0.07 & 17.7\\
  DeepFool & 0.9583 $\pm$ 0.0039 & 17.20 $\pm$ 1.48\% &  1.42 $\pm$ 0.19 & 6.0\\
  \RobustRadBS(FGSM) & 0.9585 $\pm$ 0.0034 & 13.68 $\pm$ 2.40\% & 1.60 $\pm$ 0.19 & 59.6\\
  \RobustRadBS(PGD) & \textbf{0.9649 $\pm$ 0.0028} & \textbf{13.51 $\pm$ 1.04\%} & \textbf{1.14 $\pm$ 0.08} & 9.0\\
  \RobustRadFast & 0.9574 $\pm$ 0.0042 & 16.15 $\pm$ 1.43\% & 1.53 $\pm$ 0.16 & \textbf{824.7}\\
\bottomrule
\end{tabular}
}
\caption{\textbf{Comparison of different robust radius calculation methods on CIFAR10 dataset.} The architecture is WideResNet-28 and all experiments are conducted on 2 NVIDIA RTX A5000 GPUs. Compared with standard robust radius estimation methods, our \RobustRadBS~and\RobustRadFast~accelarate much with little performance loss. Worth noting that our \RobustRadFast~achieves $\textbf{46.5}\times$ speedup compared with FAB, providing detection speed almost as fast as MSR while significant better MisD performance than MSR.}
\label{tab:RRCompare}
\end{table*}

%% file: tables/ablation_at.tex
\begin{table}[htbp]
\centering
\scalebox{0.9}{
\begin{tabular}{cccc}
\toprule
 Method & AUROC $\uparrow$ & AURC $\times$ 1000 $\downarrow$\\
\midrule
Standard & 
0.9261 $\pm$ 0.0053  & 4.08 $\pm$ 0.27 \\
 AT & 0.9548 $\pm$ 0.0069  & 1.88 $\pm$ 0.48 \\
 ReverseAT  & 0.9503 $\pm$ 0.0077 & 2.54 $\pm$ 0.61 \\
 \rat~\textbf{(Ours)} & \textbf{0.9585 $\pm$ 0.0034} & \textbf{1.60 $\pm$ 0.19} \\
\bottomrule
\end{tabular}
}
\caption{Ablation study for \rat~on CIFAR10 dataset. The architecture used is WRN-28. The confidence measure is robust radius derived by \RobustRadBS. Compared with orignial adversarial training or reversed adversarial training objective, our \rat~shows stronger ability in misclassfication detection.}
\label{tab:ATAblation}
\end{table}

%% file: sections/conclusion.tex
\label{sec:conclusion}
In this work, we studied the problem of misclassification detection (MisD). We proposed to use robust radius as a useful confidence score for MisD, which has not been explored in the field of MisD to our best knowledge. We designed two computationally efficient algorithms \RobustRadBS~and \RobustRadFast~ to estimate robust radius for MisD. Additionally, we proposed a Radius Aware Training algorithm (\rat) to boost MisD ability without additional data. Extensive experiments showed that our methods outperform baselines over various settings. Notably, compared with prior work in MisD, our methods are easier to implement, less sensitive to hyperparameter choice, does not require additional training data, and more robust to input corruptions. 

One potential limitation of our method is: \RobustRadBS~ method is shown to be more effective than the baselines, but generally, it requires more computations though it's usually affordable. Our \RobustRadFast~ accelerates the algorithm more but still brings in a slight overhead compared to MSR. We discuss this further in \cref{app:compute}.

%% file: sections/Appendix/Algorithms.tex
\section{Further discussion on \RobustRadBS~ and \RobustRadFast }
\label{app:algo}
\subsection{Algorithms}
In this section, we show the algorithms of our \RobustRadBS~ and \RobustRadFast. In \cref{alg:RR-BS}, ADV stands for any adversarial attack algorithm that generate adversarial example at given budget. In this work, we use FGSM~\cite{goodfellow2014fgsm} for its simplicity and efficiency. 
\begin{minipage}[t]{.48\textwidth} 
\SetKwComment{Comment}{\#}{}
\begin{algorithm}[H]
    \caption{\RobustRadBS: Calculating robustness radius via boundary search}\label{alg:RR-BS}
    \KwIn{$f$, $x$}
    \KwOut{Prediction robustness radius $r$}
    $r_{max} \gets r_{min}$\;
    $\AdvInput \gets \AdvAlgo(x, \ModelPred(x), r_{max})$\;
    \While {$\ModelPred(\AdvInput) = \ModelPred(x)$}{
        \tcp{Search for upper bound of $\RobustRad$}
        $r_{min} \gets r_{max}$\;
        $r_{max} \gets 2r_{max}$\;
    }
    \For{$i\gets 1 $ \KwTo $\maxiter$}{
        $r \gets (r_{min} + r_{max}) / 2$\;
        $\AdvInput \gets \AdvAlgo(x, \ModelPred(x), r)$\;
        \eIf{$\ModelPred(\AdvInput) = \ModelPred(x)$}{
            $r_{min} \gets r$\;
        }{
            $r_{max} \gets r$\;
        }
    }
\end{algorithm}
\end{minipage}
\hfill
\begin{minipage}[t]{.48\textwidth} 
\begin{algorithm}[H]
    \caption{\RobustRadFast: A fast algorithm to estimate robust radius}
    \KwIn{$f$, $x$}
    \KwOut{Prediction robust radius estimation $\FastScore$}
    Select a attacking direction: $\AdvDirection \gets sign(\nabla_x CE(x, \ModelPred)$\;
    Calculate derivative at $t=0$: 
        $\DirectionCls^\prime(0) = \nabla_\AdvDirection f(x)$\;
    Calculate $q_i$ by \cref{eq:FastSolution}\;
    $r_{\text{fast}} \gets \min_{i \neq \ModelPred} q_i$\;
    \label{alg:RRfast}
\end{algorithm}
\end{minipage}

%% file: sections/Appendix/Computation.tex
\subsection{Computational overhead}
\label{app:compute}
In this section, we discuss the computational overhead of our methods at the inference stage in terms of the number of forward and backward propagation needed for each example. For common logit-based confidence score (e.g. MSR, DOCTOR) with input preprocessing\cite{liang2017ODIN}, the calculation takes 1 forward pass and 1 backward pass.
\begin{enumerate}
    \item For \RobustRadBS, the computation cost depends on the adversarial attack method. In this work, we adopt FGSM~\cite{goodfellow2014fgsm} as the attack method, which only needs to do back-propagation once to calculate the gradient at $x$, along with $L = \UpperSearch + \maxiter + 1$ forward pass, where $\UpperSearch$ is the number of forward passes used to search the upper bound of $r$ in \cref{alg:RR-BS}. In the experiments, we impose a fixed budget $\budget = 25$ for each example and choose $\maxiter = \budget - \UpperSearch - 1$. Hence, our \RobustRadBS~ takes 1 backward pass and $\budget$ forward pass. 
    \item \RobustRadFast~ also requires one backward pass to calculate the attack direction $d$. For derivative $\DirectionCls^\prime(0)$, calculating this via finite difference approximation takes another forward pass. Thus, our \RobustRadFast~ takes 1 backward pass and 2 forward passes.
\end{enumerate} 

%% file: sections/Appendix/Experiments.tex
\section{Experimental details and additional results}
\label{app:exp}
In this section, we discuss more details of our experiments and present additional results. 

\subsection{Choice of hyperparameters for confidence scores}
As we mentioned in \cref{sec:exp}, we choose hyperparameters of each method using a validation set which contains 20\% percent of test data. We list hyperparameters of each method we tuned in \cref{tab:HyperParams}. We choose hyperparameters from $T = [0.2, 0.4, 0.6, 0.8, 1, 1.1, 1.2, 1.3, 1.4, 1.5, 2, 2.5, 3, 100, 1000]$ and $\epsilon = [0, 0.00005, 0.0001, 0.00015, 0.0002, 0.00025. 0.0003, 0.00035, 0.0004, 0.0006, 0.0008, 0.001]$ and select best hyperparameter according to AURC on the validation set. For methods with multiple hyperparameters, a grid-search on these hyperparameters is conduct to find best combinations.

For RelU\cite{gomes2023RelU}, there is another hyperparameter $\lambda$ controlling the regularization strength, which we set to $\lambda = 0.5$ following the recommendation of \cite{gomes2023RelU}. Additionally, since RelU requires additional training data for training uncertainty estimator, we split the validation set, use half of the validation data to train estimator and another half for hyperparameters tuning. 
\begin{table}[!ht]
\centering
\begin{tabular}{c|ccccccc}
Method         & MSR                  & ODIN                 & DOCTOR & RelU & \RobustRadBS & \RobustRadFast  \\ \hline
Hyperparameter & $\epsilon$ & $\epsilon, T$&   $\epsilon, T$     &   $\epsilon, T$   &  $T$  &    $T$    
\end{tabular}
\caption{Hyperparameters of each method.}
\label{tab:HyperParams}
\end{table}
\subsection{Training details of \rat}
In this section, we discuss the training details and hyperparameters used in the \rat. For all experiments, we use the SGD optimizer with a cosine annealing learning rate schedule, with 5 warmup epochs. For CIFAR10 dataset, we use a batch size of 128 to train 500 epochs from scratch and a initial learning rate of 0.2. For CIFAR100, we use a batch size of 256 to train 360 epochs from scratch and initial learning rate set to 0.1. For ImageNet dataset, we start from the pretrained weights provided by Pytorch and  use a batch size of 1024 to train for 80 epochs. We use a initial learning rate of 0.5. For all datasets, we apply Mixup~\cite{zhang2017mixup} with $\alpha=1$ for CIFAR10 and CIFAR100 and $\alpha=0.2$ for ImageNet. We choose perturbation budget $\epsilon=0.001$ for all datasets. 
\subsection{Data pre-processing}
\cite{liang2017ODIN} proposed a data pre-processing method for logit-based confidence scores:
\begin{equation}
    x^\prime = x - \epsilon \times \text{sign}[-\nabla_x \log(\ConfidScore(x)],
\end{equation}
where $\ConfidScore$ is the confidence score function and $\epsilon$ is the hyper-parameter controlling the perturbation scale. This data pre-processing is also adopted by \cite{gomes2023RelU,granese2021doctor}. We apply this pre-processing for MSR, ODIN, DOCTOR and RelU, as it's important to scores' performance.

\subsection{Understanding robust radius: mitigating the impact of overconfidence}
\begin{figure}[ht!]
\begin{subfigure}[b]{0.48\textwidth}
    \centering
    \includegraphics[scale=0.4]{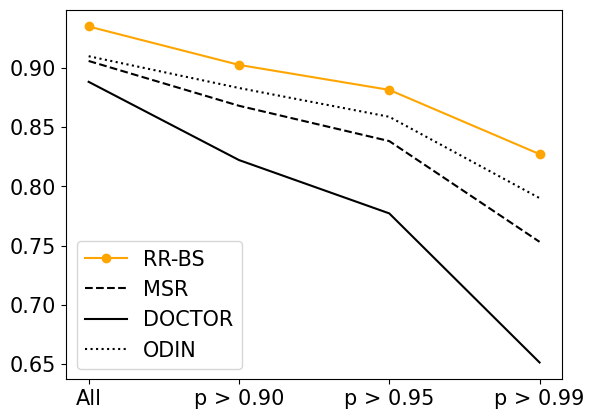}
    \caption{CIFAR10}
\end{subfigure}
\hfill
\begin{subfigure}[b]{0.48\textwidth}
    \centering
    \includegraphics[scale=0.4]{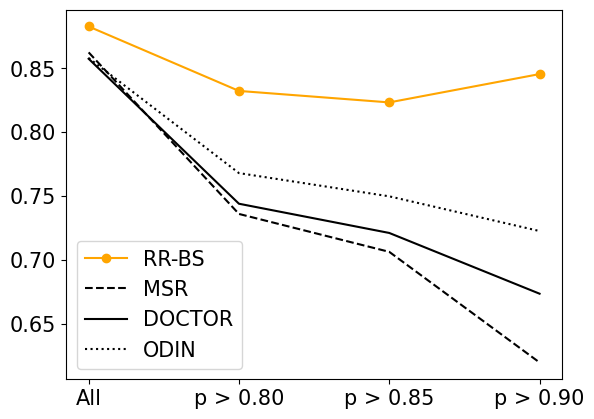}
    \caption{CIFAR100}
\end{subfigure}
\caption{\textbf{AUROC of different methods on high confidence region:} This figure shows how the performance of different methods changes when we focus on high confidence predictions. The orange line denotes our methods and the black lines denote the baselines. As shown in the figure, (1) all methods suffer different extents of performance decrease for high-confidence samples, (2) the performance of our method drops less in high-confidence region, suggesting that it has a stronger ability to detect overconfident misclassified examples.}
\label{fig:OverConfid}
\end{figure}

In this section, we study further why robust radius could outperform other confidence scores from the angle of overconfidence. 
Deep neural networks are known to have over-confident issues \cite{nguyen2015overconfident}, i.e. models may give high softmax probability even the prediction is incorrect.
Hence, a big challenge to detecting misclassified examples is to distinguish between correctly classified examples which has high confidence, and misclassified examples that have high probability due to overconfidence.
We compare the ability of different confidence scores to detect misclassified examples in the high-confidence region. 
Specifically, we filter out inputs where the model gives a predicted probability $p > \tau$ and check the discrimination ability of different methods. We show the results in \cref{fig:OverConfid}. The results suggest that (1) Distinguishing correct and incorrect examples is more challenging in high-confidence area, as the AUROC of all methods decreases when we raise the probability threshold. (2) Robust radius, however, suffers less when the prediction probability goes up. This implies robust radius is more robust to the overconfident predictions from the model, providing an explanation for the performance gain of using it as confidence measure. 

\subsection{Experiment results on CIFAR100}

\begin{table*}[htbp]
\centering
\scalebox{0.9}{
\begin{tabular}{c|c|ccccc}
\toprule
Model &  Method & AUROC $\uparrow$& FPR95 $\downarrow$ & AURC $\times$ 1000 $\downarrow$ &  Accuracy $\uparrow$& \# of extra samples $\downarrow$\\
\midrule
\multirow{9}{*}{WRN28} & MSR & 0.8665 $\pm$ 0.0014 & 52.35 $\pm$ 1.95\% & 50.40 $\pm$ 0.59 & 82.44\% & -\\
 & ODIN & 0.8730 $\pm$ 0.0030 & 45.74 $\pm$ 0.57\% & 45.44 $\pm$ 0.45 & 82.44\% & -\\
 & DOCTOR & 0.8707 $\pm$ 0.0036 & 50.74 $\pm$ 2.26\% & 47.51 $\pm$ 0.66 & 82.44\% & -\\
 & RelU & 0.8129 $\pm$ 0.0266 & 71.90 $\pm$ 8.71\% & 67.56 $\pm$ 7.28 & 82.44\% & -\\
 & OpenMix & \textbf{0.9018 $\pm$ 0.0017} & 36.16 $\pm$ 0.64\% & \textbf{34.39 $\pm$ 0.58} & \textbf{83.32\%} & 300k\\
 & Mixup & 0.8841 $\pm$ 0.0005 & 41.98 $\pm$ 0.17\% & 43.21 $\pm$ 0.35 & 82.09\% & 0\\
 & RegMixup & 0.8870 $\pm$ 0.0013 & 43.96 $\pm$ 0.55\% & 40.07 $\pm$ 0.57 & \underline{82.46\%}  & 0
  \\
 \cdashline{2-7}\noalign{\vskip 0.5ex}
 & \rat+\RobustRadBS~\textbf{(Ours)} & 0.8928 $\pm$ 0.0026 & \textbf{34.30 $\pm$ 0.88\%} & \underline{39.10 $\pm$ 0.83} & 82.14\% & 0\\
 & \rat+MSR \textbf{(Ours)} & \underline{0.8935 $\pm$ 0.0028} & \underline{36.11 $\pm$ 0.93\%} & 41.11 $\pm$ 0.87 & 82.14\%& 0 \\
 \midrule
\multirow{9}{*}{ResNet110} & MSR & 0.8462 $\pm$ 0.0020 & 55.76 $\pm$ 2.05\% & 76.65 $\pm$ 2.08 & \textbf{76.79\%} & - \\
 & ODIN & 0.8513 $\pm$ 0.0011 & 47.99 $\pm$ 1.65\% & 70.83 $\pm$ 1.43 & \textbf{76.79\%}& -\\
 & DOCTOR & 0.8500 $\pm$ 0.0011 & 49.48 $\pm$ 1.23\% & 72.19 $\pm$ 1.51 & \textbf{76.79\%}& -\\
 & RelU & 0.8144 $\pm$ 0.0052 & 61.40 $\pm$ 1.60\% & 85.49 $\pm$ 0.97 & \textbf{76.79\%} & -\\
 & OpenMix & \underline{0.8660 $\pm$ 0.0023 }& \underline{44.29 $\pm$ 0.73\%} & 71.79 $\pm$ 0.94 & 75.31\% & 300k\\
 & Mixup & 0.8479 $\pm$ 0.0015 & 51.79 $\pm$ 0.46\% & 83.93 $\pm$ 0.63 & 74.91\%& 0\\
 & RegMixup & 0.8478 $\pm$ 0.0016 & 53.08 $\pm$ 0.12\% & 74.19 $\pm$ 1.37 & \underline{76.46\%} & 0
  \\
 \cdashline{2-7}\noalign{\vskip 0.5ex}
 & \rat+\RobustRadBS~\textbf{(Ours)} & 0.8635 $\pm$ 0.0005 & \textbf{43.75 $\pm$ 0.47\%} & \textbf{67.73 $\pm$ 0.75} & 76.40\%& 0 \\
 & \rat+MSR \textbf{(Ours)} & \textbf{0.8693} $\pm$ 0.0012 & 44.72 $\pm$ 0.32\% & \underline{69.23 $\pm$ 0.30} & 76.40\%& 0 \\
\bottomrule
\end{tabular}
}
\caption{Main results for our methods and baselines on CIFAR100 dataset. Standard deviation is measures over 3 independent runs. For OpenMix~\cite{zhu2023openmix}, the OOD dataset used is 300K-Random-Image dataset~\cite{hendrycks2018deep}.}
\label{tab:MainResultsCIFAR100}
\end{table*}


%% file: sections/Appendix/RAT.tex
\section{Illustration of RAT}
\begin{figure*}[htbp]
    \centering
    \begin{subfigure}[b]{0.3\textwidth}
        \centering
        \includegraphics[width=\textwidth]{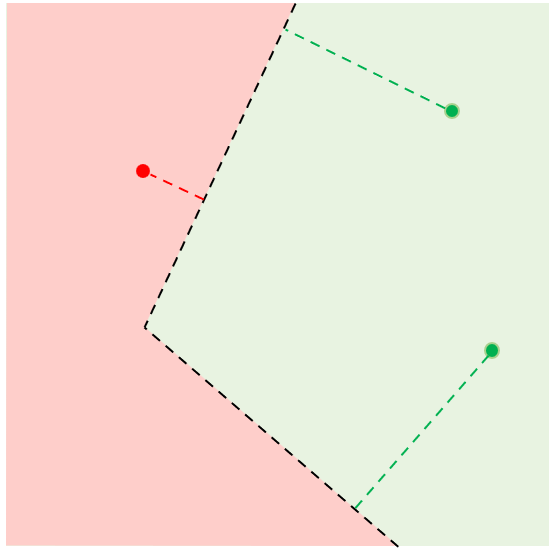}
        \caption{Standard training}
        \label{fig:sub1}
    \end{subfigure}
    \hfill
    \begin{subfigure}[b]{0.3\textwidth}
        \centering
        \includegraphics[width=\textwidth]{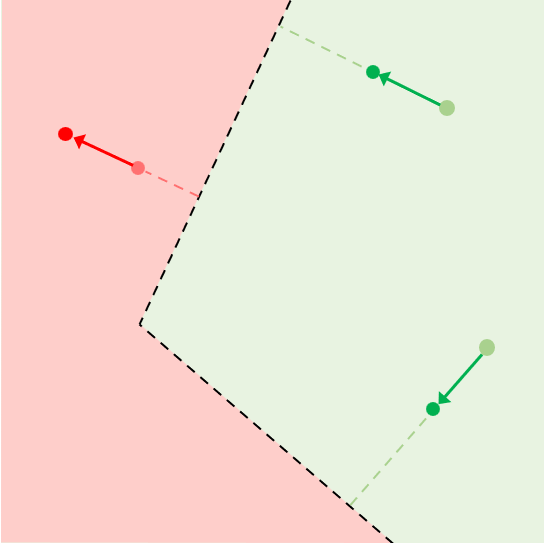}
        \caption{Adversarial Training}
        \label{fig:sub2}
    \end{subfigure}
    \hfill
    \begin{subfigure}[b]{0.3\textwidth}
        \centering
        \includegraphics[width=\textwidth]{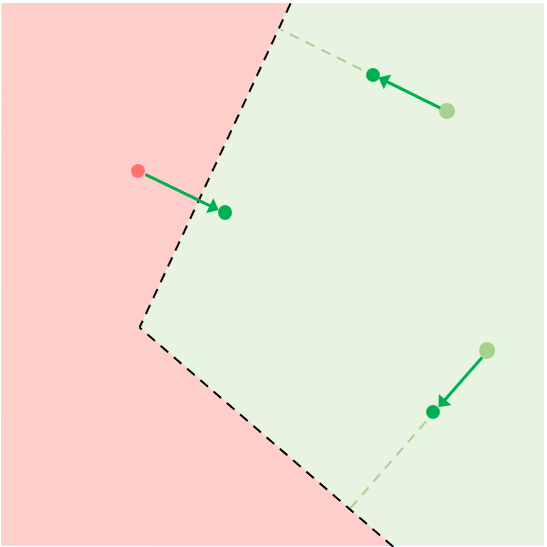}
        \caption{Our \rat.}
        \label{fig:sub3}
    \end{subfigure}
    
    \caption{Illustration of different training methods. (Left) For standard model, we observe that correct examples locate further from boundary compared to misclassified ones, verifying common belief. (Middle) Common adversarial training methods create adversarial example by pushing them towards ``incorrect region" (Right) In our \rat, we differ from standard adversarial training by pulling incorrect examples to "correct region". If the incorrect example is close to "correct region", the perturbed example will be correctly classified. This encourages the model to avoid "big mistake", i.e. misclassified examples that are far from decision boundary.}
    \label{fig:method_comp}
\end{figure*}